%% file: main.tex
\documentclass[conference]{IEEEtran}
\IEEEoverridecommandlockouts
\usepackage{cite}
\usepackage{amsmath,amssymb,amsfonts}
\usepackage{algorithmic}
\usepackage{graphicx}
\usepackage{textcomp}
\usepackage{xcolor}
\usepackage{multirow}

\usepackage[ruled,linesnumbered]{algorithm2e}

\def\BibTeX{{\rm B\kern-.05em{\sc i\kern-.025em b}\kern-.08em
    T\kern-.1667em\lower.7ex\hbox{E}\kern-.125emX}}
\begin{document}

\title{Image classification network enhancement methods based on knowledge injection}

\input{Sections/00-Author/main}

\maketitle

\input{Sections/00-Abstract/main}

\input{Sections/01-Keywords/main}

\input{Sections/02-Introduction/main}

\input{Sections/03-RelatedWork/main}

\input{Sections/04-Algorithm/main}

\input{Sections/05-Dataset/main}

\input{Sections/06-Evaluation/main}

\input{Sections/07-Conclusion/main}

\input{Sections/08-Acknowledgment/main}

\input{Sections/09-Reference/main}
\end{document}

%% file: Sections/00-Author/main.tex
\author{
\IEEEauthorblockN{1\textsuperscript{st} Yishuang Tian}
\IEEEauthorblockA{\textit{School of Software Engineering} \\
\textit{Xidian University}\\
Xi'an, China \\
tyshuang@mail.hfut.edu.cn}
\and
\IEEEauthorblockN{2\textsuperscript{nd} Ning Wang}
\IEEEauthorblockA{\textit{School of Software Engineering} \\
\textit{Xidian University}\\
Xi'an, China \\
ningwang@stu.xidian.edu.cn}
\and
\IEEEauthorblockN{3\textsuperscript{rd} Liang Zhang}
\IEEEauthorblockA{\textit{School of Software Engineering} \\
\textit{Xidian University}\\
Xi'an, China \\
liangzhang@xidian.edu.cn}
}

%% file: Sections/00-Abstract/main.tex
\begin{abstract}
The current deep neural network algorithm still stays in the end-to-end training supervision method like Image-Label pairs, which makes traditional algorithm is difficult to explain the reason for the results, and the prediction logic is difficult to understand and analyze. The current algorithm does not use the existing human knowledge information, which makes the model not in line with the human cognition model and makes the model not suitable for human use. In order to solve the above problems, the present invention provides a deep neural network training method based on the human knowledge, which uses the human cognition model to construct the deep neural network training model, and uses the existing human knowledge information to construct the deep neural network training model. This paper proposes a multi-level hierarchical deep learning algorithm, which is composed of multi-level hierarchical deep neural network architecture and multi-level hierarchical deep learning framework. The experimental results show that the proposed algorithm can effectively explain the hidden information of the neural network. The goal of our study is to improve the interpretability of deep neural networks (DNNs) by providing an analysis of the impact of knowledge injection on the classification task. We constructed a knowledge injection dataset with matching knowledge data and image classification data. The knowledge injection dataset is the benchmark dataset for the experiments in the paper. Our model expresses the improvement in interpretability and classification task performance of hidden layers at different scales.
\end{abstract}
%The code and dataset can be obtained at the following link.
%The current deep neural network training method still stays in the end-to-end training supervision method, the explanation of the hidden tensor space in the neural network is very lacking, and the training data of the deep neural network is single, and the current algorithm does not use the existing human knowledge information. The end-to-end training method and single training data make it impossible for humans to analyze the internals of the neural network, which making the model a "black box model" and humans cannot understand the hidden information and logic. Therefore, in order to solve the above problems, we proposed the algorithm "knowledge injection" in this paper. Our method uses knowledge graphs and natural language as explanation media to inject neural networks hierarchically, and our algorithm utilizes knowledge information of different scales and integrates these knowledge so that the model can obtain rich knowledge at multiple scales , to improve the interpretability of the model.
%并且我们构建了带有知识数据与图像分类数据相匹配的知识注入数据集作为基准数据集进行了论文中的实验。我们的模型表达出了不同尺度下的隐藏层可解释能力与分类任务性能上的提高。我们的代码与数据集可以在以下链接获取。

%% file: Sections/01-Keywords/main.tex

%% file: Sections/02-Introduction/main.tex
\section{Introduction}

\begin{figure}[htbp]
\centerline{\includegraphics[width=1.0\linewidth]{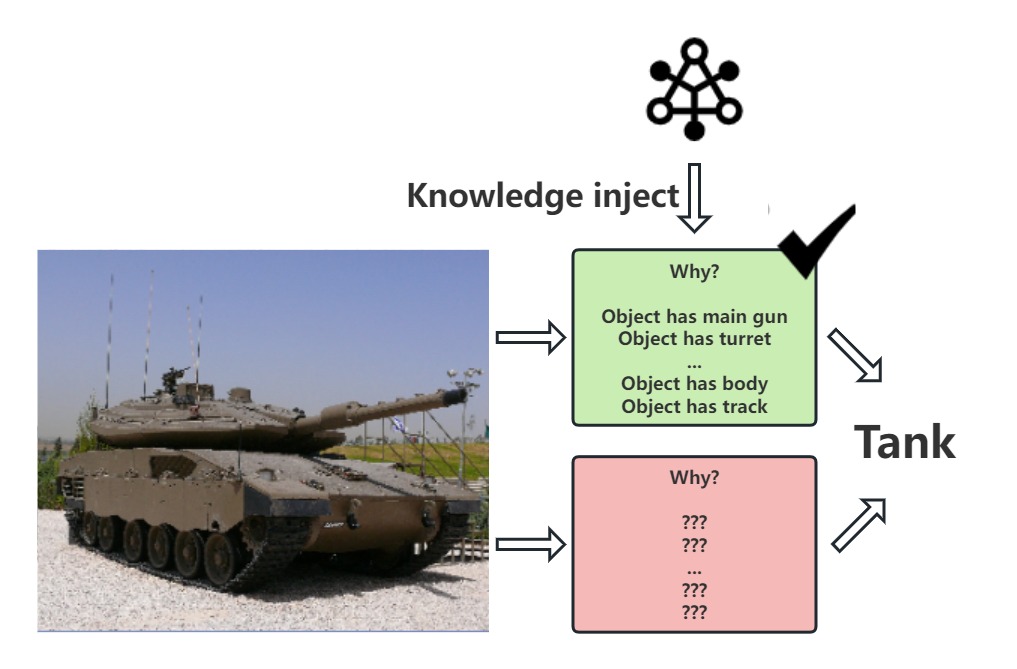}}
%传统神经网络由于其黑盒的特性导致无法表达模型内部隐藏层的预测行为逻辑，使用知识注入则可以优化传统神经网络在隐藏层可解释性差的缺点。
\caption{Traditional neural networks due to its black box characteristic cannot express the logical behavior of the internal hidden layer of the model, using knowledge injection can optimize the disadvantages of traditional neural networks in poor interpretability of hidden layers.}
\label{02-Introduction/figure-1}
\end{figure}

Image classification is a very important research direction in the field of computer vision, and the feature extraction network is used in the field of computer vision for object detection, semantic segmentation, and multimodal model. Therefore, developing excellent image classification networks in the field of computer vision is very important

\begin{figure*}[htbp]
\centerline{\includegraphics[width=1.0\linewidth]{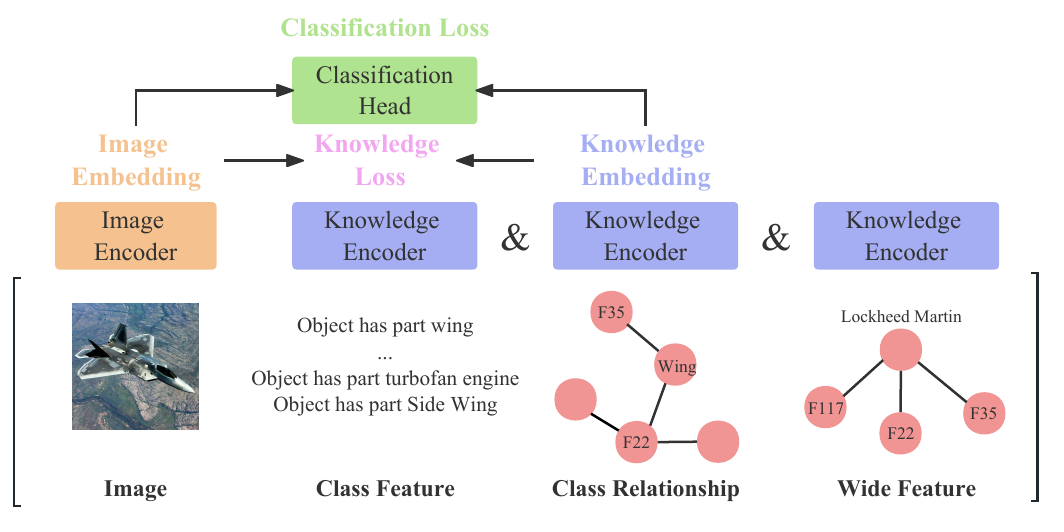}}
%我们引入多尺度的知识数据加入到训练数据中去以丰富模型的语义部件理解能力与空间理解能力，我们分别采用了三元组形式的类别特征、类别间部件特征与类别关系特征，将这三种不同尺度下的知识信息进行监督的到知识损失进行优化。
\caption{We introduce multi-level knowledge data to enrich the semantic part of the model and the spatial understanding ability of the model, respectively. We use three-way feature of category, 1) category-level feature, 2) category part relationship feature and 3) category relationship feature to add different level of knowledge information to the knowledge loss during training, and optimize the knowledge loss.}
\label{02-Introduction/figure-2}
\end{figure*}

And in the field of deep learning, the problem of explaining the model's hidden layer has not been effectively solved, and the reliability and trustworthiness of the black box network have been questioned in many application scenarios, such as medical and autonomous driving, because if the model is not trustworthy, it will cause very big problems. So the explainability of neural networks is what many researchers are striving for.

The majority of neural networks do not have semantic logical relationships cognition, which will lead to the model unable to explain the reasons for the choice of judgments, and people are unable to control the neural network to learn the semantic component of the logic of human beings. As shown in Figure 1, the model can distinguish the image belongs to the tank category, but cannot explain the reasons for the selection of tanks, so we combine the human knowledge to train the neural network to try to let the model can learn the semantic logical relationship of human beings to make a judgement output.

We use natural language to carry the description information of the object to control the learning components of the model as the starting point of the neural network logic, and then use the relationship knowledge graph between the parts of the object to guide the logical reasoning of the previous step, and finally use the knowledge graph with a larger scale to give the model a wider understanding of generalization ability.

The proposed algorithm consists of two parts: the first part is the knowledge injection process, which is divided into three stages: 1) extracting the knowledge graph and natural language information from the knowledge base; 2) converting the knowledge graph and natural language information into a format that the neural network can understand; 3) injecting the knowledge graph and natural language information into the neural network. The second part is the training classification process of the neural network. In this paper, we injected the knowledge graph and natural language information into the neural network by the above algorithm and trained the neural network to obtain the knowledge graph and natural language explanation of the neural network.

%图像分类任务是计算机视觉领域内非常重要的研究方向，分类任务中的特征提取网络被用在计算机视觉领域中的目标检测、语义分割、多模态等领域，所以开发优秀的图像分类网络在计算机视觉领域是非常重要的。并且在深度学习领域，对模型的隐藏层解释问题一直没有被有效地解决，这种黑盒网络的可靠性、可信任性在许多应用场景下遭到了质疑，如医疗、自动驾驶等行业，因为如果模型不是可信任的则会引发非常大的问题。所以神经网络的可解释也是许多研究者所为之努力的。
%大多数的神经网络没有语义逻辑关系的认知，这会导致模型无法解释做出选择判断的原因，人们也无法控制神经网络对人类的语义部件逻辑进行学习。如图1模型可以分辨出图像属于坦克类别但是无法解释做出坦克选择的原因，所以我们结合人类的知识来对神经网络进行训练来尝试让模型可以学习得到人类的语义逻辑关系来进行判别输出。
%在本文中我们利用自然语言来携带物体的描述信息来控制模型学习部件特征作为神经网络逻辑的起点，再使用物体的部件之间的关系知识图谱来指导前一步的逻辑推理，最后使用更广泛的知识图谱来赋予模型更大尺度上的理解泛化能力。
%

%% file: Sections/03-RelatedWork/main.tex
\section{Related Works}

\begin{figure*}[ht]
\centerline{\includegraphics[width=1.0\linewidth]{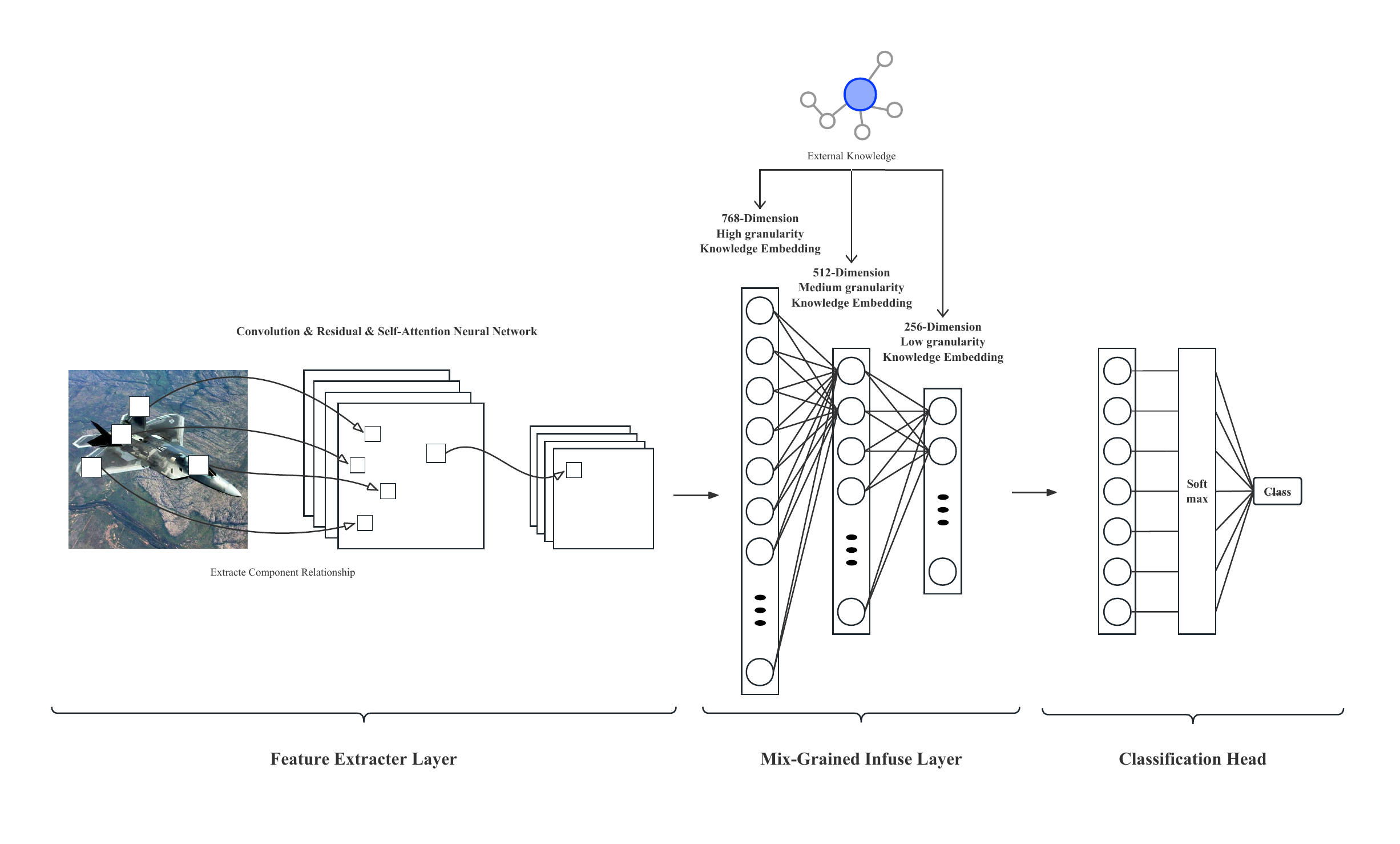}}
%模型示意图，模型由三个主要模块组成部分组成，分别为适配不同骨干网络（ResNet，ViT等）的特征提取层、混合知识注入层与分类头模块。
\caption{The model consists of three main modules, namely the feature extraction layer, the mix-grained infuse layer, and the classification head module. 1) The feature extraction layer is composed of different ResNet, ViT, etc. modules to adapt to different backbone networks; 2) The mixing knowledge injection layer is composed of activation function layers and fully connected layers; 3) The classification head module is composed of a fully connected layer.}
\label{04-Algorithm/figure-1}
\end{figure*}

\subsection{Interpretable network}
%在可解释网络领域上，先驱者们做出了非常优秀的算法来解决神经网络的可解释问题，其中有使用逆向梯度来求出模型的激活图的算法Grad-CAM，这种方法在黑盒训练完成后对模型进行注意力权重分析来解释模型所关注的部分，也有使用掩模来遮住图像中不同部分再让神经网络进行预测以解释神经网络的算法LIME，除此之外也有很多其它可解释算法，但是这些可解释算法大多数为进行黑盒训练后再进行可解释分析，并没有在训练的过程中进行可解释监督，并且这些可解释算法也没有使用到更加丰富的人类语义逻辑知识来对模型本体进行增强。
In the explainable machine learning domain, pioneers have made excellent algorithms to solve the explainable problem of neural networks, among which Grad-CAM, which uses reverse gradient to obtain the model's activation map, is a method that performs attention weight analysis on the model after completing black box training to explain the model's attention, and LIME, which uses masks to cover different parts of the image and then let the neural network make predictions to explain the neural network algorithm, and many other explainable algorithms, but most of these explainable algorithms do not perform explainable supervision during training, and these explainable algorithms have not used more abundant human language semantics to enhance the model itself.
\subsection{Multimodal model}
%多模态模型是目前计算机视觉与自然语言处理研究者们的重要研究方向，多模态领域涌现了如CLIP、BLIP、GPT4等优秀的多模态模型，这些模型使用图像特征提取器与文本特征提取器对图像文本对进行向量对齐监督，得到一个对图像以及文本特征有着非常强大的提取能力的多模态，这些多模态模型可以应用在Zero-Shot分类、图像内容文本生成、VGQ等一系列任务，并且在下游任务表现了非常强的能力。但是目前的多模态模型仍然处于图像文本多模态模型，使用文本作为一个特征与图像特征进行匹配对齐，这些算法没有利用到更加丰富的人类知识模态。
Multi-modal models are a current research direction in computer vision and natural language processing researchers, and the multi-modal domain has emerged excellent multi-modal models such as CLIP, BLIP, and GPT4, which use image features extractors and text features extractors to align the vector of images and texts, and obtain a multi-modal model with very powerful feature extraction capabilities on image and text. These multi-modal models can be applied to Zero-Shot classification, image content text generation, VQA(Visual Question Answering), and other tasks, and the downstream tasks have very strong capabilities. However, the current multi-modal models still rely on text as a feature to match and align the image feature, which does not fully utilize the more abundant human knowledge modality.
\subsection{Knowledge Embedding}
%在知识的表示任务中，前人总结了非常多的方法进行知识图谱的嵌入表示，提取知识图谱中的知识特征为向量，比如有做节点嵌入的算法DeepWalk，该算法在图中的每个节点上进行随机游走生成多个随机游走序列，使用该序列中的节点临近关系作为嵌入依据来对节点进行嵌入，也有使用图卷积网络GCN进行图自编码器的嵌入算法VGAE，VGAE将经典的自编码器架构Encoder-Decoder思想用到了图谱的嵌入中，这些方法都可以将图进行表示，将图中的语义特征进行提取表示。
In the task of knowledge representation, many methods have been summarized to represent knowledge graphs, such as extracting the knowledge features of knowledge graphs as vectors, such as the DeepWalk algorithm that extracts the knowledge features of knowledge graphs as vectors by randomly walking on the nodes in the knowledge graph, and the VGAE algorithm that uses the classic Encoder-Decoder architecture to embed the knowledge graph, these methods can be used to represent the image, to extract the semantic features of the image.
%\subsection{Knowledge Injection}

%% file: Sections/04-Algorithm/main.tex
\section{Algorithm}

%我们的算法主要分为三个部分，第一个部分着重于多尺度知识的表示嵌入，第二个部分重点在知识注入网络的模型设计结构，第三个部分在于知识注入网络的优化方法。这三个部分是明显区别于传统深度神经网络的训练过程，所以我将在这三方面进行详细说明。
We can divide our algorithm into three parts. The first part focuses on the representation of multi-scale knowledge embedding. The second part focuses on the model design structure of knowledge injection into the network. The third part is about the optimization method of knowledge injection algorithm. These three parts are obviously different from the training process of traditional deep neural networks, so I will explain in detail in these three aspects.

\subsection{Knowledge Embedding}

%我们所使用到的知识数据需要通过知识嵌入算法进行处理后才可以在模型的训练中进行使用，我们给模型注入了三个不同尺度下的知识信息，这些尺度分别为类别特征尺度，类别间关系尺度，广泛特征尺度，这些知识信息可以携带丰富的人类语义信息来让深度神经网络的推导过程变得更加语义化与逻辑化。接下来我将分别介绍这三个知识嵌入方法。
We use knowledge data to be processed by knowledge embedding algorithms before they can be used in model training. We injected three different scale of knowledge information into the model, which are the category feature scale, the category relationship scale, and the wide feature scale. These knowledge information can carry rich human semantic information to make the deep neural network's inference process more semantic and logical. Then I will introduce these three knowledge embedding methods.

\subsubsection{Class Feature Embedding}

%在单一类别中，我们为了让张量汲取更多的类别特征信息，我们需要对类别三元组知识进行预处理后再进行编码。我们的知识的基础形式为三元组，即主题关系客体的形式，我们若需要将三元组组成一个含有语义的自然语言，则需要将其预处理为句子再使用语言模型进行编码。我们这里的做法使用句子模板来进行处理，由于我们的三元组均为主语谓语宾语的关系，所以我们可以新建一个句子模板，将对应的实体与关系进行填充即刻创建自然语言形式的知识数据，再使用语言模型BERT对其进行编码得到最终的类别特征编码。
In a single category, we need to preprocess the category triples to extract more category feature information. Our knowledge base is in the form of triples, i.e., theme-relation-object, if we need to form a natural language containing semantics from the triple, we need to preprocess it into a sentence and use a language model to encode it. Our approach uses sentence templates to process it. Since all of our triples are in the form of subject-verb-object, we can create a new sentence template, and fill in the corresponding entities and relations to create natural language-form knowledge data, and then use the language model BERT to encode the final category feature encoding.

\subsubsection{Class Relationship Embedding}

%当我们在一个比物体特征的尺度更大的尺度上时，我们会注意到一个物体的部件关系，这个物体是由那些部件进行组成，这些部件相互之间的空间位置关系这些更加宏观的知识信息。在我们的算法中为了将这种知识信息进行表示，我们使用到了图嵌入算法，将一个物体的不同部件看作图数据结构中的一个节点，将部件与部件之间的关系看作图数据结构中的一个边值，这样就可以将含有相同部件的类别进行联系，将所有的类别之间的关系进行表示整理，最终我们将这个图进行节点嵌入，即可生成一个蕴含类别间关系的知识嵌入。
When we observe an object at a scale larger than its physical characteristics, we will notice the relationship between the parts of an object, this object is composed of these parts, the spatial relationship between the parts of these more macroscopic knowledge information. In our algorithm, in order to express this knowledge information, we use the embedding algorithm of the graph, the different parts of an object are regarded as nodes of the graph data structure, the relationship between the parts of the parts is regarded as a value of the edge of the graph data structure, which can be connected with the same parts of the category, and the relationship between all the categories can be expressed and organized, and finally, we can generate a knowledge embedding that contains the relationship between the categories.

\input{Sections/04-Algorithm/algorithm-1}

\subsubsection{Wide Feature Embedding}

%在类别间关系嵌入过程后，我们将知识认知尺度更加放大，我们可以关注到标签类别外的更多知识关系信息，这里的知识已经脱离训练使用的基本数据集，而是从互联网上动态搜寻广泛的知识信息。我们这里使用的广泛知识使用数据集类别作为根节点，在维基百科中进行搜索，得到与根节点相关的所有类别信息，这里将爬取得到的一张更广泛的知识图谱进行节点嵌入则可表示更广泛的知识信息。
After the embedding process between categories, we will enlarge the knowledge cognition scale, and we can pay attention to more knowledge relationship information outside the tag category, here the knowledge has been separated from the basic data set used for training, but is dynamically searched from the Internet for a wide range of knowledge information. We use a wide range of knowledge data set as the root node category in Wikipedia and search for all related category information. Here, we will crawl the wide range of knowledge graph and embed it into the node to represent more extensive knowledge information.

\subsection{Injection Module}

%我们提出了注入模块作为知识注入算法的基础模块，该模块的作用是在高维张量空间对知识嵌入进行监督，并且在不同尺度下的知识嵌入之间进行转换或下采样。模块由多个线性层、激活函数以及Dropout层所构成，其中线性层与激活函数可以使得知识张量可以在高维空间进行非线性变换进行不同表示空间下的转换，Dropout层可以让模型在训练过程中降低过拟合现象发生的概率。将知识注入模块加入传统网络即得到算法中所使用的知识注入模型。
We propose to use an embedding module as the basic module of knowledge injection algorithm, whose function is to supervise the knowledge injection in high-dimensional tensor space, and to transform or downsample the knowledge injection in different scales. The module is composed of multiple linear layers, activation functions, and Dropout layers, where the linear layers and the activation functions can make the knowledge tensor transform nonlinearly in different representations space, and the Dropout layers can reduce the probability of overfitting in the training process. When the knowledge injection module is added to the traditional network, the knowledge injection model is obtained.

\subsection{Optimization Method}

\subsubsection{Knowledge Optimization}

\input{Sections/04-Algorithm/algorithm-2}

%在模型的优化过程中，第一个步骤是使用知识嵌入对特征提取网络进行优化，这个步骤将不包括分类头MLP的特征提取层参数进行优化，使用多个尺度的知识嵌入借助知识注入模块对神经网络中的参数进行优化，这个步骤为广义上的模型预训练，可以让特征提取网络与知识特征进行匹配。这里的优化使用的监督方式参考了对比学习的监督方式，因为是特征张量之间的匹配对齐任务，我们引入正负样本的监督机制以及余弦相似度作为损失的计算方法来进行监督。
In the optimization process of the model, the first step is to use knowledge embedding to optimize the feature extraction network of the classifier MLP, this step will not include the optimization of the feature extraction layer parameters of the classifier MLP, using multiple scales of knowledge embedding assisted by the knowledge injection module to optimize the parameters of the neural network, this step is generally speaking to pre-train the model, it can make the feature extraction network match the knowledge features. The optimization of this step is based on the supervised learning method of the contrast learning method, because it is a matching task between the feature vectors, we introduce the supervised mechanism of positive and negative samples and the calculation method of the cosine similarity as the loss function to supervise.

\subsubsection{Classification Optimization}

\input{Sections/04-Algorithm/algorithm-3}

%当我们完成知识优化阶段后，我们冻结特征提取网络中的所有网络参数，只优化分类头MLP中的参数，这样可以保证模型隐藏层的隐藏层解释能力以及最终模型的分类能力，这里的监督优化方式则为传统的交叉熵损失。
After we complete the knowledge optimization stage, we freeze the feature extraction network's all network parameters and optimize only the parameter of the classification head MLP, so as to ensure the interpretation ability of the hidden layer of the hidden layer of the model and the classification ability of the final model. The supervised optimization method is the traditional cross-entropy loss.

%% file: Sections/04-Algorithm/algorithm-1.tex
\begin{algorithm}[!ht]
  \SetAlgoLined
  \KwData{
  knowledge graph G(V,E)\;
  windows size $W$\;
  embedding size $d$\;
  walks per vertex $\gamma$\;
  walk length $t$}
  
  \KwResult{matrix of vertex representations $\Phi \in \mathbb{R}^{|V| \times b}$}

  Initialization: Sample $\Phi$ from $\mathcal{U}^{|V| \times b}$\;
  Build a binary Tree $T$ from $V$\;
  \While{i=0 to $\gamma$}{
    $\mathcal{O}$ = Shuffle($V$)\;
    \While{$v_i \in \mathcal{O}$}{
      $W_{v_i}$ = RandomWalk($G$, $v_i$, $t$)\;
      SkipGram($\Phi$,$W_{v_i}$,$W$)\;
    }
  }
  \caption{Relationship Embedding}
  \label{algo:1-algorithm}
\end{algorithm}

%% file: Sections/04-Algorithm/algorithm-2.tex
\begin{algorithm}[!ht]
  \SetAlgoLined
  \KwData{
  image set $S$\;
  knowledge epoch $e$\;
  knowledge tensor $\xi$\;
  }
  \While{0 to $e$}{
    \While{$I$ in $S$}{
      $\nu$ = $Forward_{Hidden}$ ($I$)\;
      $Loss = -\frac{\sum_{i=1}^{n} \nu_{i} \times \xi_{i}}{\sqrt{\sum_{i=1}^{n}\nu_{i}^{2}} \times \sqrt{\sum_{i=1}^{n}\xi_{i}^{2}}}$\;
      $Backward(Loss)$\;
    }
  }
  \caption{Knowledge Optimization}
  \label{algo:2-algorithm}
\end{algorithm}

%% file: Sections/04-Algorithm/algorithm-3.tex
\begin{algorithm}[!ht]
  \SetAlgoLined
  \KwData{
  image set $S$\;
  image label set $L$\;
  classfication epoch $e$\;
  }
  \While{0 to $e$}{
    \While{$I,l$ in $zip(S, L)$}{
      $\nu$ = $Forward_{Hidden}$ ($I$)\;
      $\iota$ = $Forward_{MLP}$ ($\nu$)\;
      $Loss = CrossEntropy(\iota, l)$\;
      $Backward(Loss)$\;
    }
  }
  \caption{Classification Optimization}
  \label{algo:3-algorithm}
\end{algorithm}

%% file: Sections/05-Dataset/main.tex
\input{Sections/05-Dataset/02-table}

\section{Knowledge-Image Classification Dataset}

\input{Sections/05-Dataset/01-table}

%我们为了将知识注入算法进行实验测试，我们收集整理了一个知识注入数据集，该数据集收集了99类共计3000个图像知识图谱对，知识图谱使用三元组的形式进行储存，里面记录着实体-关系-实体的数据。
We collected and organized a knowledge injection data set for experimenting to inject knowledge into algorithms. The data set collected 99 kinds of knowledge graph, with 3000 triples. The knowledge graph is stored in the form of three-tuple, which records the entity-relationship-entity.

\subsection{Knowledge Dataset}

%如下表中的数据所示，我们将一个目标根据人类的先验知识拆分为不同的部件，再将这些不同的部件进行特征的描述，并且将部件与部件之间的关系进行描述，最终形成了知识数据集中的多条三元组数据。这些三元组可以组成一个知识图谱来对数据集中的目标类别特征进行精确精细地描述，这些描述包括部件组成、空间关系、部件特征等。
The data in the following table shows that we will divide a target into different parts according to human prior knowledge, describe the features of the different parts, and describe the relationships between the parts, finally forming a multi-set of three-way data in the knowledge data set. These three-way sets can be used to describe the precise and fine characteristics of the target category in the knowledge data set in a more accurate and fine manner, which include the composition of parts, spatial relationships, part features, etc.

\subsection{Image Dataset}

%我们将图像数据集范围定在了军事武器这个范围内，并且将目标分为了海陆空三个大类别，在这三个大类别中再进行细分得到共计99类的图像数据，这些图像数据都有对应的知识数据作为匹配，并且均图像均有明显的特征像素。图像数据集的整理方式与传统的图像分类数据集相似，可以将其单独提取作为传统图像分类任务的数据集。
We will focus on the military weapon data set in the image data set, and divide the target into three major categories, namely sea, land and air, and divide the three major categories into a total of 99 subcategories. These image data sets all have corresponding knowledge data as a matching, and all have obvious features. The data set organization method is similar to that of the traditional image classification data set, and it can be extracted separately as a traditional image classification task data set.

%% file: Sections/05-Dataset/02-table.tex
\begin{table*}[ht]
\caption{Dataset Content}
\begin{center}
\begin{tabular}{|c|c|c|}
\hline
\textbf{Class Name}&\textbf{Image Data}&\textbf{Knowledge Data} \\
\hline
Kiev Aircraft carrier&\begin{minipage}{0.5\columnwidth}\raisebox{-0.5\height}{\includegraphics[width=\linewidth]{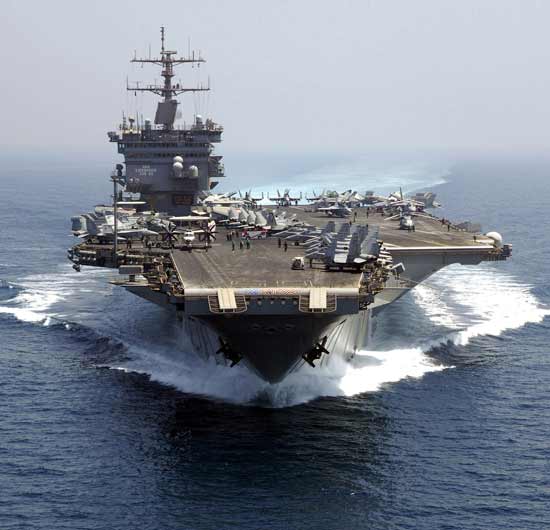}}\end{minipage}&\begin{minipage}{0.7\columnwidth}
Kiev Aircraft carrier has/part deck \par
Kiev Aircraft carrier has/part cabin \par
Kiev Aircraft carrier has/part shipbridge \par
Kiev Aircraft carrier has/part radar \par
Kiev Aircraft carrier has/part runway \par
Kiev Aircraft carrier has/part shipboard aircraft \par
deck is/top/of cabin \par
Shipbridge is/top/of deck \par
radar is/top/of Shipbridge \par
runway is/top/of deck \par
radar is/top/of Shipbridge \par
runway is/top/of deck \par
shipboard aircraft is/top/of deck \par
$\ldots$ \par
deck is/adjacent/to runway \par
shipbridge is/adjacent/to radar
\end{minipage}\\
\hline
PP-19&\begin{minipage}{0.5\columnwidth}\raisebox{-0.5\height}{\includegraphics[width=\linewidth]{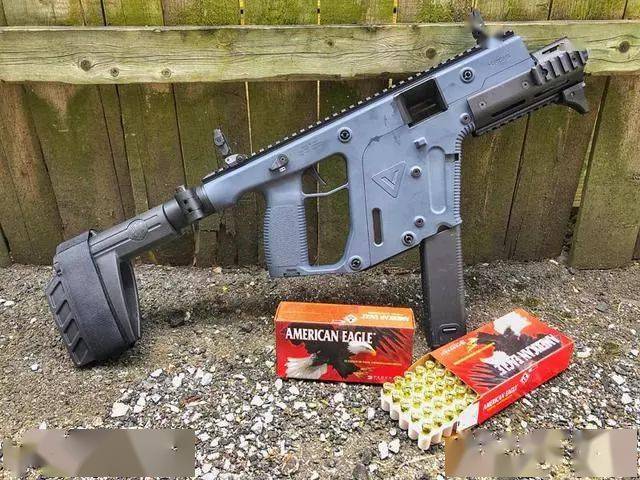}}\end{minipage}&\begin{minipage}{0.7\columnwidth}
PP-19 has/part barrel \par
PP-19 has/part trigger \par
PP-19 has/part gun barrel \par
PP-19 has/part Tubular magazine \par
PP-19 has/part grip \par
PP-19 has/part buttstock \par
barrel is/right/of gun barrel \par
buttstock is/left/of gun barrel \par
grip is/below/of gun barrel \par
trigger is/below/of gun barrel \par
$\ldots$ \par
grip is/adjacent/to trigger \par
trigger is/adjacent/to Tubular magazine
\end{minipage}\\
\hline
OH-58&\begin{minipage}{0.5\columnwidth}\raisebox{-0.5\height}{\includegraphics[width=\linewidth]{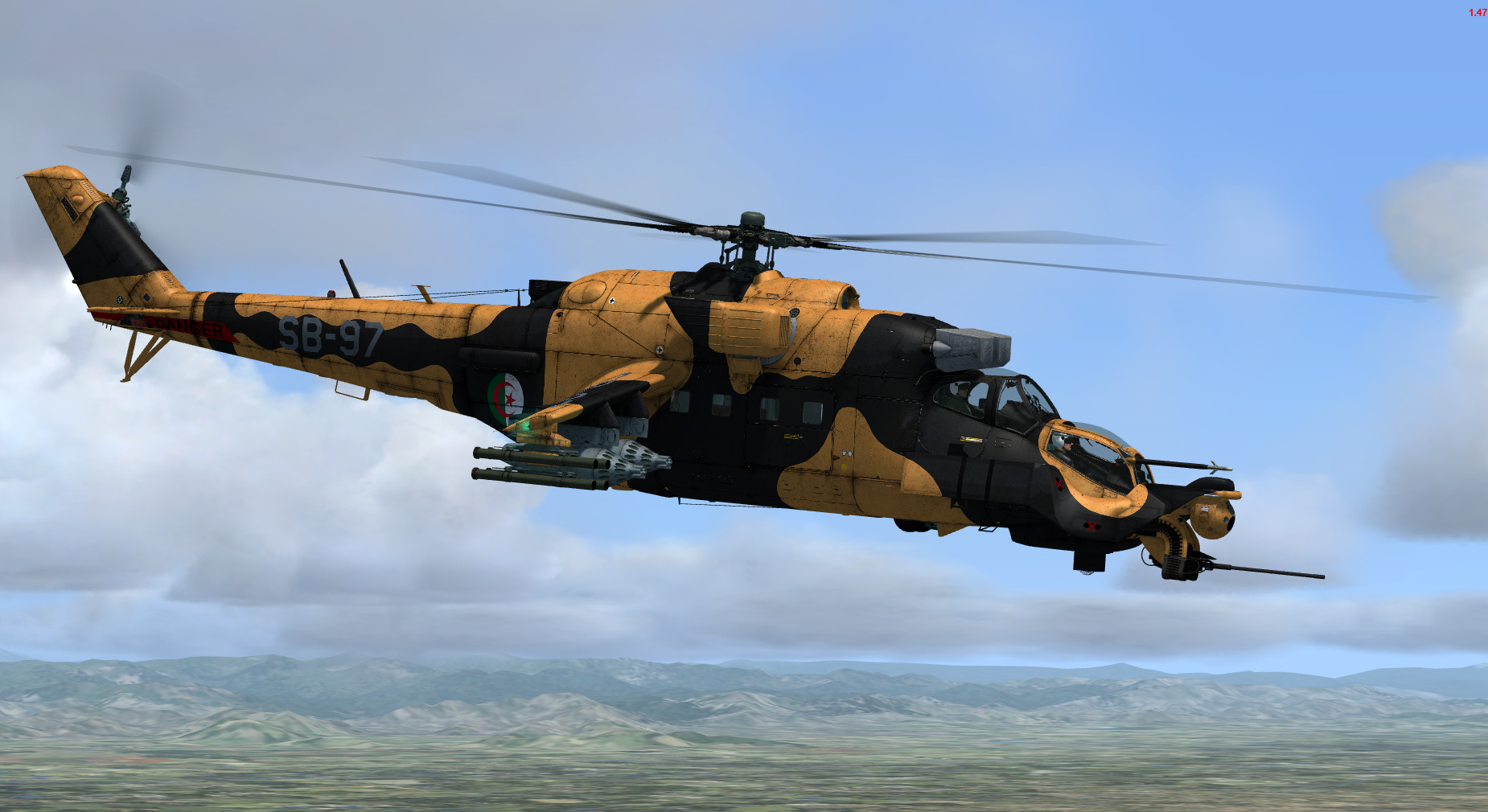}}\end{minipage}&\begin{minipage}{0.7\columnwidth}
OH-58 has/part nose \par
OH-58 has/part fuselage \par
OH-58 has/part Tail \par
OH-58 has/part rotor \par
OH-58 has/part Tail Rotor \par
OH-58 has/part Sled landing gear \par
$\ldots$ \par
fuselage is/adjacent/to Tail \par
nose is/adjacent/to Sled landing gear
\end{minipage}\\
\hline
F/A-18E/F&\begin{minipage}{0.5\columnwidth}\raisebox{-0.5\height}{\includegraphics[width=\linewidth]{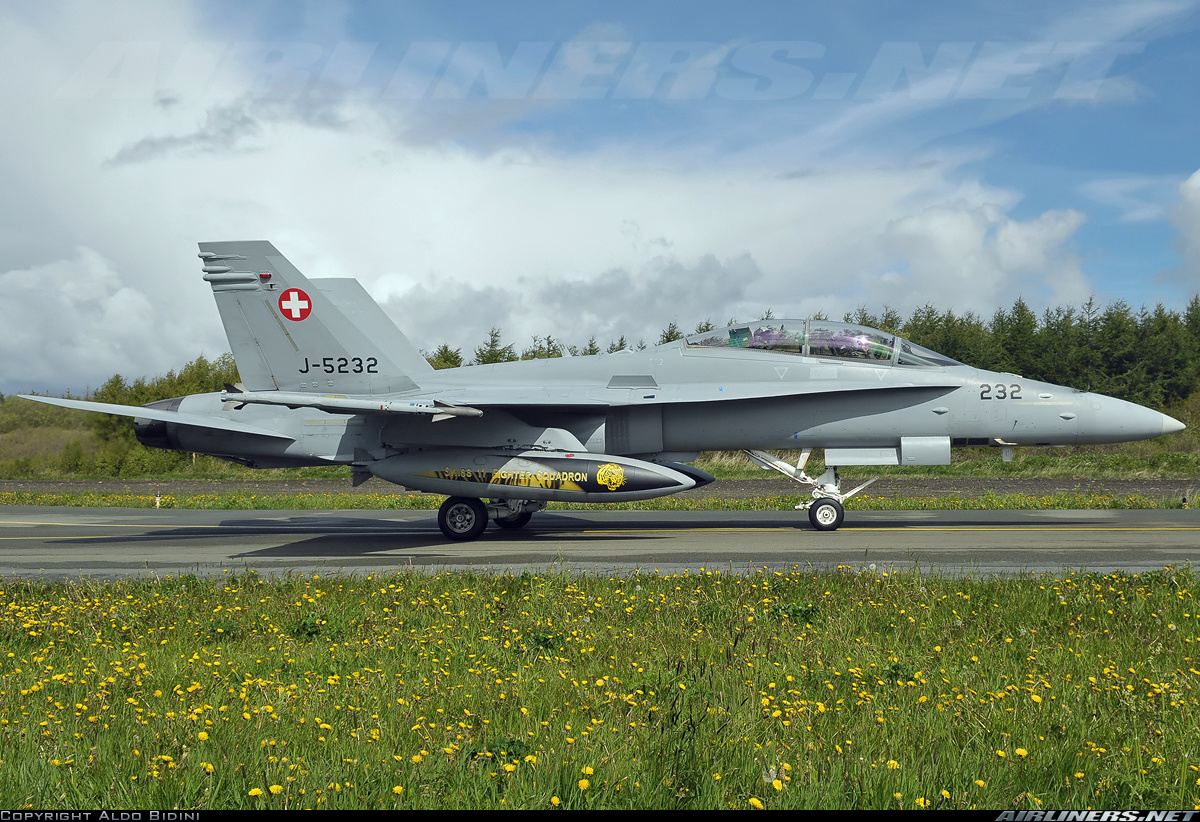}}\end{minipage}&\begin{minipage}{0.7\columnwidth}
FA-18E-F has/part nose \par
FA-18E-F has/part fuselage \par
FA-18E-F has/part wing \par
FA-18E-F has/part Sideband wing \par
FA-18E-F has/part Tail \par
FA-18E-F has/part Turbofan engine \par
FA-18E-F has/part Wheeled landing gear \par
nose is/front/of fuselage \par
wing is/middle/of fuselage \par
Sideband wing is/front/of wing \par
$\ldots$ \par
fuselage is/adjacent/to Turbofan engine \par
Tail is/adjacent/to Turbofan engine
\end{minipage}\\
\hline
\end{tabular}
\label{Dataset-Table}
\end{center}
\end{table*}

%% file: Sections/05-Dataset/01-table.tex
\begin{table}[htbp]
\caption{Data Quantity}
\begin{center}
\begin{tabular}{|c|c|c|}
\hline
\multicolumn{2}{|c|}{\textbf{Dataset}}&\textbf{Number} \\
\hline
\multirow{2}{*}{Knowledge Data} & Training Set & 10,000 Triplets \\
\cline{2-3}
& Validating Set & 1,500 Triplet \\
\hline
\multirow{2}{*}{Image Data} & Training Set & 70,000 Images \\
\cline{2-3}
& Validating Set & 10,500 Images \\
\hline
\end{tabular}
\label{Dataset-Table}
\end{center}
\end{table}

%% file: Sections/06-Evaluation/main.tex
\section{Evaluation}

%为了评价知识注入算法在实际的性能上的表现，我们使用了三种方法对知识注入算法的优势进行了评价，这三个方法分别为正确率指标、Grad-CAM注意力分析与隐藏层的解释效果。其中隐藏层解释这个特性是我们的算法所带来的独特模型可解释方法的体现。
To evaluate the performance of knowledge injection algorithm in real-world, we used three methods to evaluate the advantages of knowledge injection algorithm, which are accuracy evaluation indicator, Grad-CAM attention analysis, and hidden layer interpretation effect. Among them, the hidden layer interpretation is the unique model interpretation method brought by our algorithm.

\subsection{Accuracy}

%首先是分类模型中非常重要的正确率指标，我们为了展示出我们算法的能力，我们设计了多个消融实验对模型进行了测试。算法中使用了三个不同尺度的知识作为训练数据所以我们在这三个尺度上分别做了消融实验，我们使用了残差网络的代表网络ResNet与Transformer的代表网络ViT，每种分别选择了三个网络进行了消融实验，记录了这六个网络在验证数据集上的损失以及正确率指标。得到的结果如下表所示，可以看到相同骨干网络下在经过知识注入后的验证集损失与正确率指标有了部分提升，这部分提升来自于外部的人类知识来引导构建模型的语义与逻辑推导能力，由此让模型获得了性能的提升。

\input{Sections/06-Evaluation/table}

First, we will focus on the important accuracy metric in the classification model, to show the ability of our algorithm, we designed a series of classification experiments. The algorithm uses three different scales of knowledge as training data, so we did three classification experiments on three different scales, and we used the representative network ResNet and ViT of the Transformer as the representative network, and selected three networks for each classification experiment. The results of the six networks on the validation dataset are shown in the table below, which can be seen that the loss of the same backbone network is improved after knowledge injection, and the accuracy is improved to some extent, which comes from the external human knowledge to guide the construction of model semantics and logical inference ability, thus making the model performance improved.

\subsection{Grad-CAM}

%为了可以使用更加直观的方法对模型的语义部件理解能力的提升进行展示，我们使用了Grad-CAM算法计算模型对分类目标的热力图，观察热力图的高权重部分是否集中于目标本体上来反映知识注入算法给模型带来的影响。如下表所示，可以看到使用知识注入算法得到的模型相较于传统模型可以让注意力权重更加集中于所需要识别的目标本体，这样就可以反映出模型在目标特征识别能力上的提升。
To demonstrate the improvement of the semantic understanding ability of the model for the use of more intuitive methods, we used Grad-CAM algorithm to calculate the heat map of the classification target of the model, and observed the high-weight parts of the heat map to reflect the influence of knowledge injection algorithm on the model. As shown in the following table, it can be seen that the model obtained by knowledge injection can be concentrated on the target body of the target, which can reflect the improvement of the target recognition ability of the model.

\input{Sections/06-Evaluation/GradCAM}

\subsection{Hidden Layer Explanation}

%由于在隐藏层使用知识对模型进行了监督，所以知识注入算法也给模型带来了隐藏层的解释能力，知识嵌入可以将含有相同特征的类别进行提取，在知识注入算法下视觉网络也保留了这种相同特征归纳的能力，知识注入网络可以在隐藏层就对识别目标进行张量维度上的解释，神经网络可以将含有相似的类别进行高维聚类，将不同的特征在不同的围度进行表示，这样可以使用降维算法明显看出神经网络内部的特征层分辨行为，我们将这些特征向量进行降维即可得到如下表效果。
Knowledge injection algorithms can help neural networks interpret the hidden layer by using knowledge for model supervision. Knowledge injection can extract classes with similar features in the hidden layer, and the visual network can retain the ability to group classes with similar features. Knowledge injection networks can explain the classification target in the tensor form of the hidden layer, and neural networks can group classes with similar features in different dimensions. This can make it easier to distinguish the features in the hidden layer using dimensionality reduction. We can get the following table effect.

\input{Sections/06-Evaluation/HiddenLayer}

%% file: Sections/06-Evaluation/table.tex
\begin{table}[htbp]
\caption{Ablation Analysis}
\begin{center}
\begin{tabular}{|c|c|c|c|c|}
\hline
\textbf{Vision}&\multicolumn{3}{|c|}{\textbf{Knowledge Inject}}&\textbf{Valid} \\
\cline{2-4} 
\textbf{Backbone}&\textbf{\textit{KI-S}}&\textbf{\textit{KI-M}}&\textbf{\textit{KI-L}}&\textbf{Accuracy} \\
\hline
& $\times$ & $\times$ & $\times$  & 71.3\% \\
\cline{2-5}
& $\times$ & $\times$ & \textcolor{green}{\checkmark}  & 73.8\% \\
\cline{2-5}
ResNet-18& $\times$ & \textcolor{green}{\checkmark} & $\times$  & 71.9\% \\
\cline{2-5}
& \textcolor{green}{\checkmark} & $\times$ & $\times$  & 70.1\% \\
\cline{2-5}
& \textcolor{green}{\checkmark} & \textcolor{green}{\checkmark} & \textcolor{green}{\checkmark}  & 74.1\% \\
\hline
& $\times$ & $\times$ & $\times$  & 74.7\% \\
\cline{2-5}
& $\times$ & $\times$ & \textcolor{green}{\checkmark}  & 76.9\% \\
\cline{2-5}
ResNet-50& $\times$ & \textcolor{green}{\checkmark} & $\times$  & 76.1\% \\
\cline{2-5}
& \textcolor{green}{\checkmark} & $\times$ & $\times$  & 74.9\% \\
\cline{2-5}
& \textcolor{green}{\checkmark} & \textcolor{green}{\checkmark} & \textcolor{green}{\checkmark}  & 77.5\% \\
\hline
& $\times$ & $\times$ & $\times$  & 75.3\% \\
\cline{2-5}
& $\times$ & $\times$ & \textcolor{green}{\checkmark}  & 77.2\% \\
\cline{2-5}
ResNet-152& $\times$ & \textcolor{green}{\checkmark} & $\times$  & 76.6\% \\
\cline{2-5}
& \textcolor{green}{\checkmark} & $\times$ & $\times$  & 75.9\% \\
\cline{2-5}
& \textcolor{green}{\checkmark} & \textcolor{green}{\checkmark} & \textcolor{green}{\checkmark}  & 78.1\% \\
\hline
& $\times$ & $\times$ & $\times$  & 68.5\% \\
\cline{2-5}
& $\times$ & $\times$ & \textcolor{green}{\checkmark}  & 69.1\% \\
\cline{2-5}
ViT-B/32& $\times$ & \textcolor{green}{\checkmark} & $\times$  & 69.7\% \\
\cline{2-5}
& \textcolor{green}{\checkmark} & $\times$ & $\times$  & 68.0\% \\
\cline{2-5}
& \textcolor{green}{\checkmark} & \textcolor{green}{\checkmark} & \textcolor{green}{\checkmark}  & 69.6\% \\
\hline
& $\times$ & $\times$ & $\times$  & 75.2\% \\
\cline{2-5}
& $\times$ & $\times$ & \textcolor{green}{\checkmark}  & 76.3\% \\
\cline{2-5}
ViT-B/16& $\times$ & \textcolor{green}{\checkmark} & $\times$  & 75.7\% \\
\cline{2-5}
& \textcolor{green}{\checkmark} & $\times$ & $\times$  & 75.5\% \\
\cline{2-5}
& \textcolor{green}{\checkmark} & \textcolor{green}{\checkmark} & \textcolor{green}{\checkmark}  & 76.5\% \\
\hline
& $\times$ & $\times$ & $\times$  & 78.7\% \\
\cline{2-5}
& $\times$ & $\times$ & \textcolor{green}{\checkmark}  & 80.5\% \\
\cline{2-5}
ViT-L/16& $\times$ & \textcolor{green}{\checkmark} & $\times$  & 79.6\% \\
\cline{2-5}
& \textcolor{green}{\checkmark} & $\times$ & $\times$  & 79.1\% \\
\cline{2-5}
& \textcolor{green}{\checkmark} & \textcolor{green}{\checkmark} & \textcolor{green}{\checkmark}  & 81.1\% \\
\hline
\end{tabular}
\label{tab1}
\end{center}
\end{table}

%% file: Sections/06-Evaluation/GradCAM.tex
\begin{table}[ht]
\caption{Grad-CAM Analysis}
\begin{center}
\begin{tabular}{|c|c|}
\hline
\textbf{Before Injection}&\textbf{After Injection} \\
\hline
\begin{minipage}{0.4\columnwidth}\raisebox{-0.5\height}{\includegraphics[width=\linewidth]{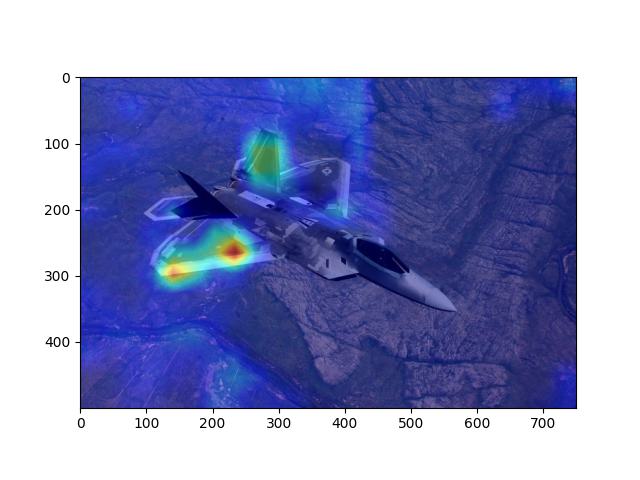}}\end{minipage}&\begin{minipage}{0.4\columnwidth}\raisebox{-0.5\height}{\includegraphics[width=\linewidth]{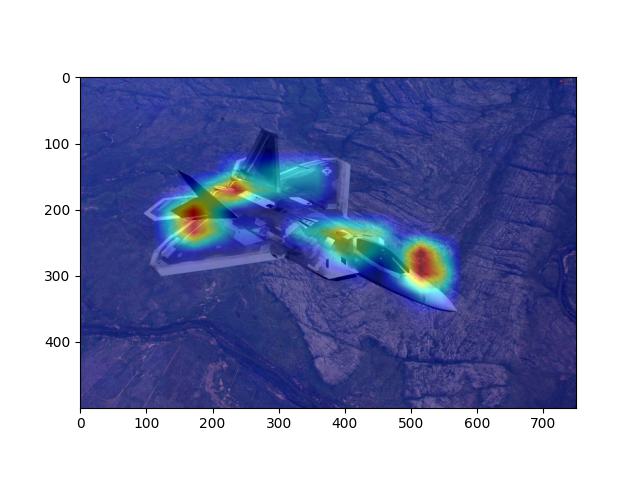}}\end{minipage}\\
\hline
\begin{minipage}{0.4\columnwidth}\raisebox{-0.5\height}{\includegraphics[width=\linewidth]{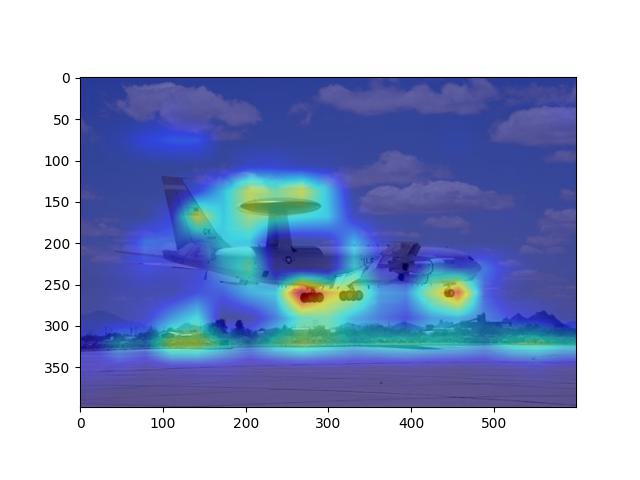}}\end{minipage}&\begin{minipage}{0.4\columnwidth}\raisebox{-0.5\height}{\includegraphics[width=\linewidth]{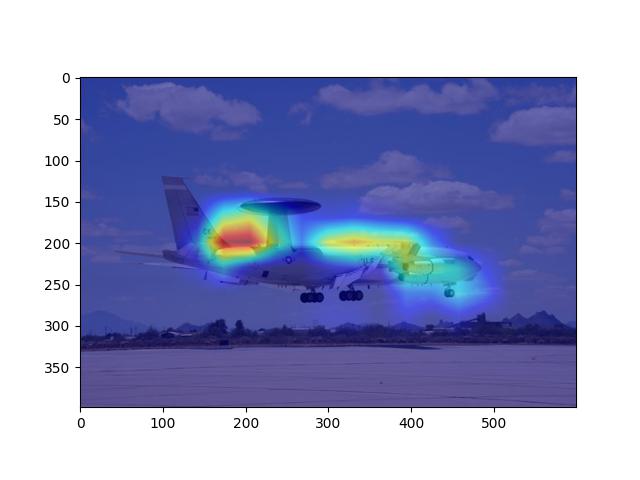}}\end{minipage}\\
\hline
\begin{minipage}{0.4\columnwidth}\raisebox{-0.5\height}{\includegraphics[width=\linewidth]{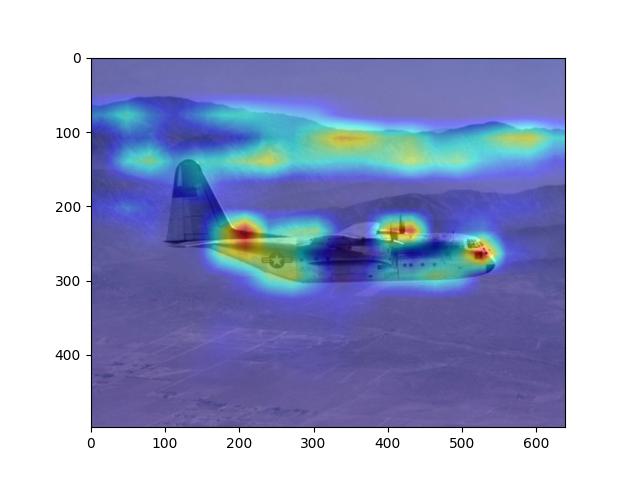}}\end{minipage}&\begin{minipage}{0.4\columnwidth}\raisebox{-0.5\height}{\includegraphics[width=\linewidth]{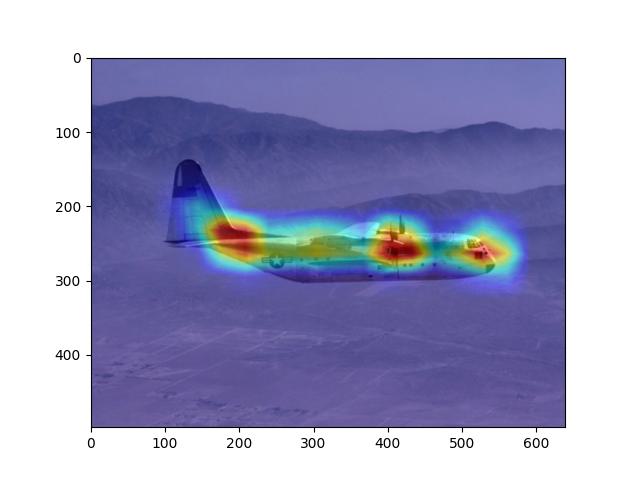}}\end{minipage}\\
\hline
\begin{minipage}{0.4\columnwidth}\raisebox{-0.5\height}{\includegraphics[width=\linewidth]{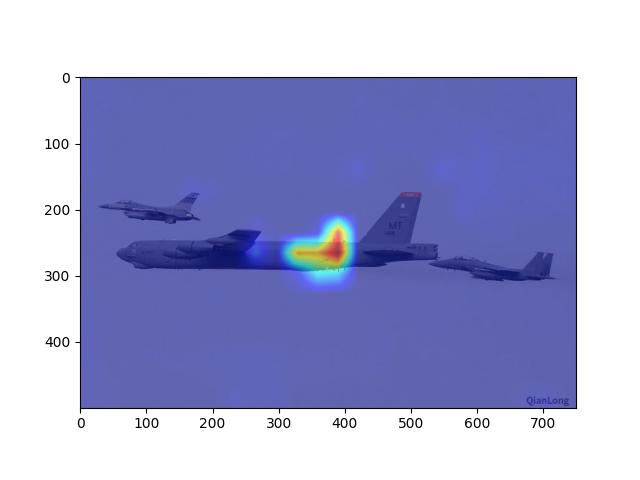}}\end{minipage}&\begin{minipage}{0.4\columnwidth}\raisebox{-0.5\height}{\includegraphics[width=\linewidth]{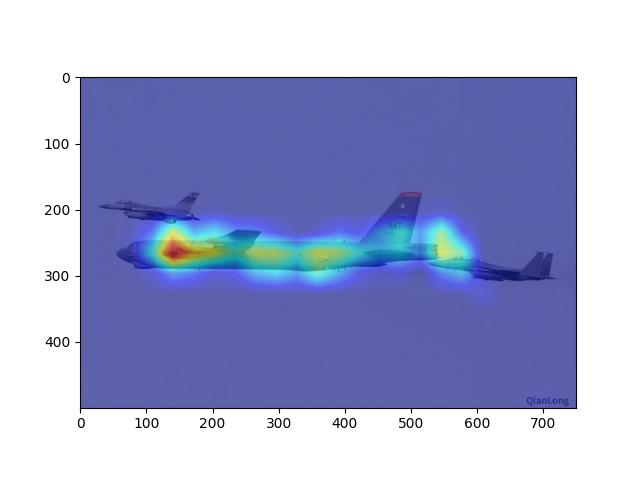}}\end{minipage}\\
\hline
\end{tabular}
\label{Dataset-Table}
\end{center}
\end{table}

%% file: Sections/06-Evaluation/HiddenLayer.tex
\begin{table}[ht]
\caption{Hidden Layer Explanation}
\begin{center}
\begin{tabular}{|c|c|}
\hline
\textbf{Original Image}&\textbf{Hidden Layer Explanation} \\
\hline
\begin{minipage}{0.4\columnwidth}\raisebox{-0.5\height}{\includegraphics[width=\linewidth]{Sections/05-Dataset/KievAircraftcarrier.jpeg}}\end{minipage}&\begin{minipage}{0.4\columnwidth}\raisebox{-0.5\height}{\includegraphics[width=\linewidth]{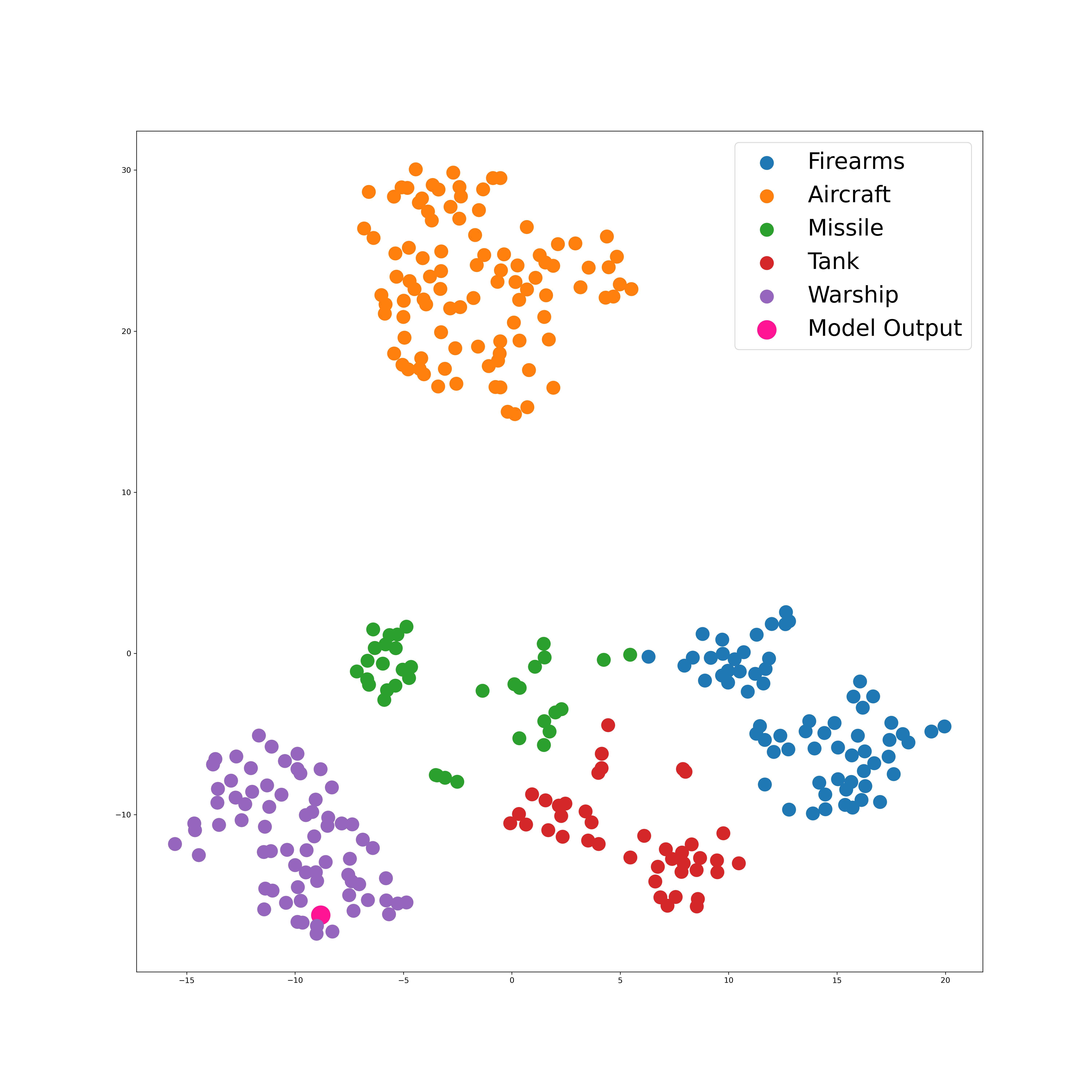}}\end{minipage}\\
\hline
\begin{minipage}{0.4\columnwidth}\raisebox{-0.5\height}{\includegraphics[width=\linewidth]{Sections/05-Dataset/PP-19.jpeg}}\end{minipage}&\begin{minipage}{0.4\columnwidth}\raisebox{-0.5\height}{\includegraphics[width=\linewidth]{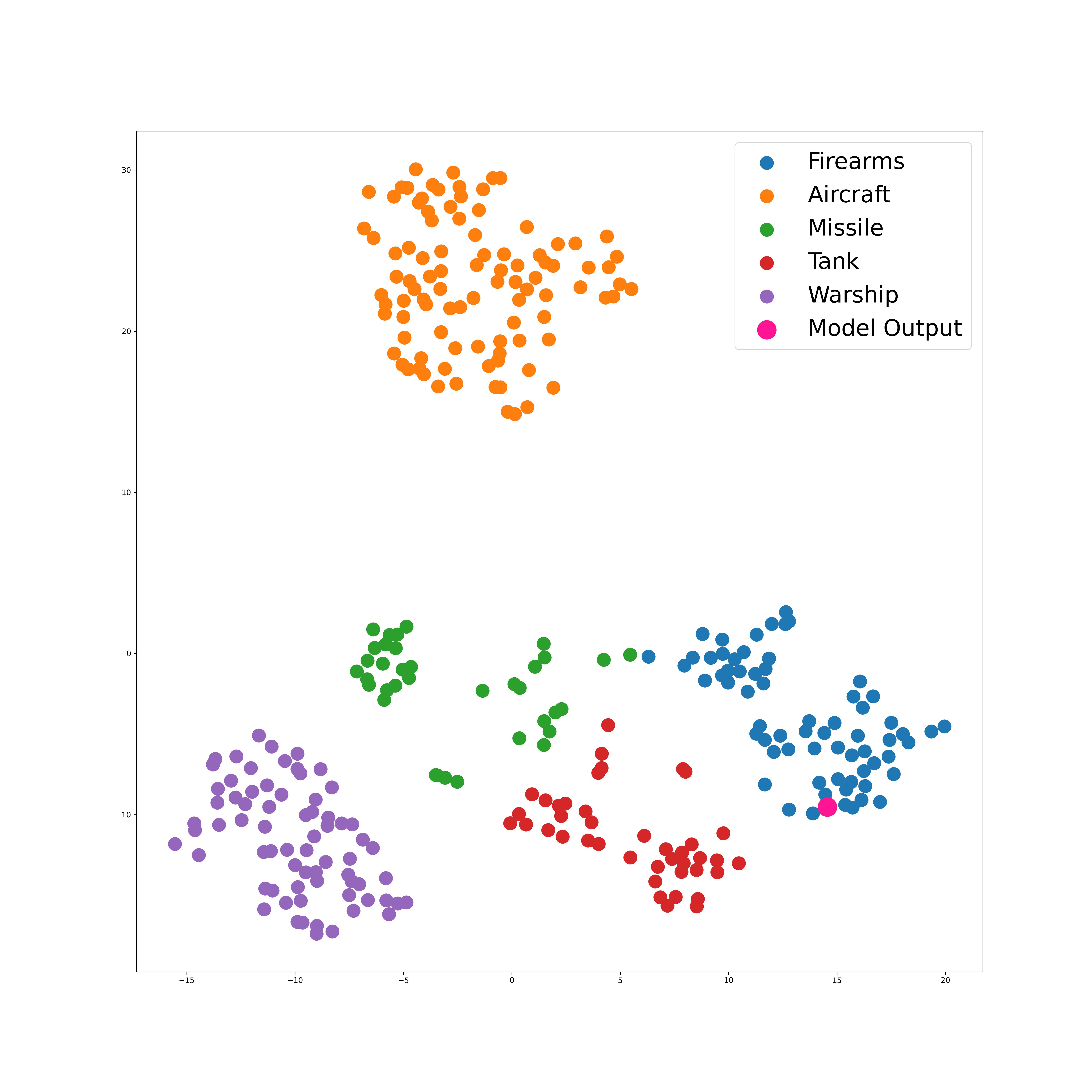}}\end{minipage}\\
\hline
\begin{minipage}{0.4\columnwidth}\raisebox{-0.5\height}{\includegraphics[width=\linewidth]{Sections/05-Dataset/OH-58.jpeg}}\end{minipage}&\begin{minipage}{0.4\columnwidth}\raisebox{-0.5\height}{\includegraphics[width=\linewidth]{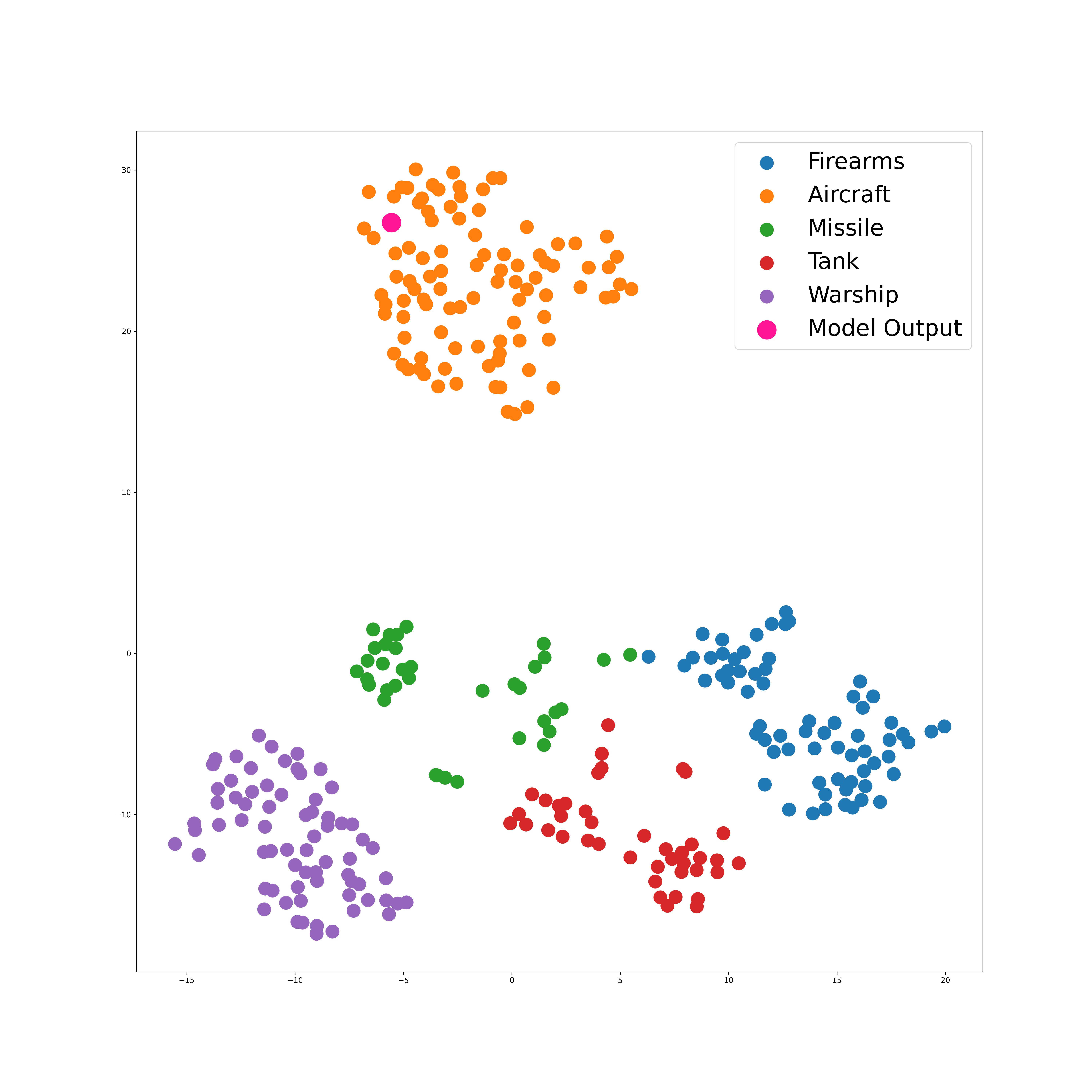}}\end{minipage}\\
\hline
\begin{minipage}{0.4\columnwidth}\raisebox{-0.5\height}{\includegraphics[width=\linewidth]{Sections/05-Dataset/FA18EF.jpeg}}\end{minipage}&\begin{minipage}{0.4\columnwidth}\raisebox{-0.5\height}{\includegraphics[width=\linewidth]{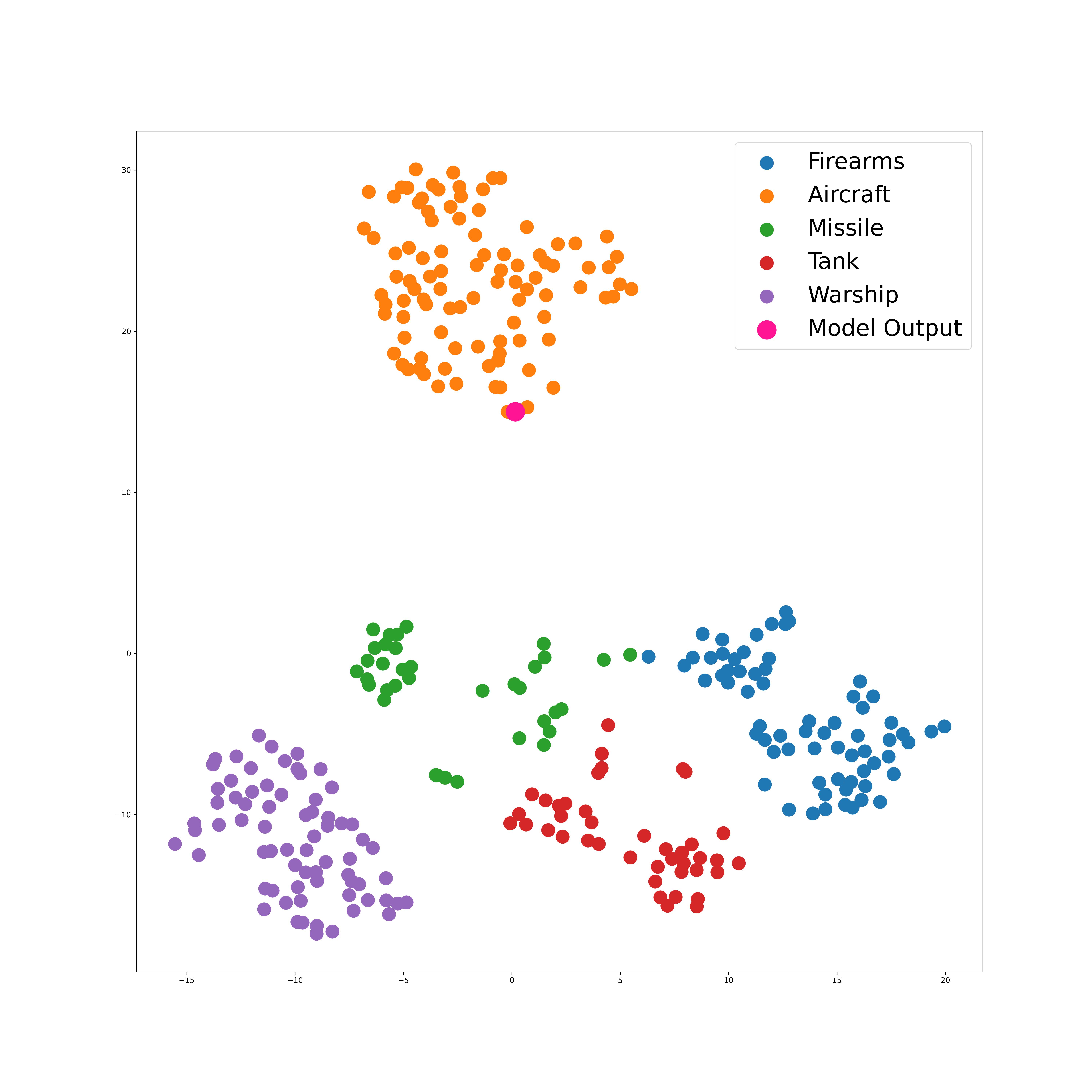}}\end{minipage}\\
\hline
\end{tabular}
\label{Dataset-Table}
\end{center}
\end{table}

%% file: Sections/07-Conclusion/main.tex
\section{Conclusion}
%在本文中我们提出了使用知识作为外部先验知识对图像分类网络进行增强的算法，并且公开了知识注入数据集以供使用。具体上来说，我们开发了知识注入层网络，提出了知识注入算法，算法使用了自然语言以及知识图谱形式的人类先验知识分层级地在神经网络的训练过程中进行了监督增强，并开发了多尺度知识融合算法来统筹不同尺度下的知识表示并且依此进行监督训练以解决神经网络的不可解释性，最终使得模型可以在多个隐藏层以及分类结果上可以有着很好的可解释表现与分类性能的提高。
In this paper, we propose using knowledge as the external prior knowledge for image classification networks to enhance the network, and release the knowledge injection dataset for using. Specifically, we develop the knowledge injection layer network, propose the knowledge injection algorithm, the algorithm uses the natural language and knowledge graph form human prior knowledge in several levels to the training process of the neural network for the supervision enhancement, and develop the multi-level knowledge fusion algorithm to aggregate the knowledge representation of different levels and train the network for the supervision enhancement, to solve the unexplainedness of the neural network, and finally make the model achieve the good explainability and classification performance on several hidden layers and classification results.

%% file: Sections/08-Acknowledgment/main.tex
\section*{Acknowledgment}

Thanks Xidian University for giving us a platform to finish this work. Thanks to every researcher who participated in this paper. Everyone has completed a series of work such as data sets, algorithms, and training. Without them, there would be no this paper. I also sincerely thank the sponsors who sponsored this research, and we have completed this paper together.